\documentclass[11pt,a4paper]{article}
\usepackage[hyperref]{acl2019}
\usepackage{times}
\usepackage{latexsym}
\usepackage{url}
\usepackage{graphicx}
\usepackage{fancybox}
\usepackage[lined,boxed,ruled]{algorithm2e}

\aclfinalcopy 

\newcommand{\name}{\textsc{PerspectroScope}}
\newcommand{\datasetname}{\textsc{Perspectrum}}

\definecolor{DarkRed}{RGB}{130,25,0}
\definecolor{DarkGreen}{RGB}{30,130,30}

\newcommand{\ignore}[1]{}
\newcommand{\claim}{\emph{claim}}
\newcommand{\perspective}{\emph{perspective}}
\newcommand{\evidence}{\emph{evidence}}

\newcommand{\set}[1]{\{#1\}}

\title{
\name: A Window to the World of Diverse Perspectives 
}

\author{Sihao Chen, Daniel Khashabi, Chris Callison-Burch, Dan Roth \\
University of Pennsylvania \\
{\tt \small \{sihaoc,danielkh,ccb,danroth\}@cis.upenn.edu}
}

\date{ }

\begin{document}
\maketitle
\begin{abstract}

This work presents \name, a web-based system which lets users query a discussion-worthy natural language claim, and extract and visualize various perspectives in support or against the claim,
along with evidence supporting each perspective. 
The system thus lets users explore various perspectives that could touch upon aspects of the issue at hand.
The system is built as a combination of retrieval engines and learned textual-entailment-like classifiers built using  a few recent developments in natural language understanding. 
To make the system more adaptive, expand its coverage, and improve its decisions over time,
our platform employs various mechanisms to get corrections from the users.


\name\ is available at 
{\small\url{github.com/CogComp/perspectroscope}}.\footnote{A brief demo of the system: \url{https://www.youtube.com/watch?v=MXBTR1Sp3Bs}. }

\end{abstract}

\section{Introduction}
One key consequence of the information revolution is a significant increase and a contamination of our information supply. The practice of fact-checking won't suffice to eliminate the biases in text data we observe, as the degree of factuality alone does not determine whether biases exist in the spectrum of opinions visible to us. To better understand controversial issues, one needs to view them from a diverse yet comprehensive set of {\em perspectives}.

Understanding most nontrivial {\em claim}s requires insights from various \perspective s.
Today, we make use of search engines or recommendation systems to retrieve information relevant to a claim, but this process carries multiple forms of \emph{bias}. In particular, they are optimized relative to the claim (query) presented, and the popularity of the relevant documents returned, rather than with respect to the diversity of the \perspective s presented in them or whether they are supported by evidence.


While it might be impractical to show an exhaustive spectrum of views with respect to a \emph{claim}, cherry-picking a small but diverse set of \emph{perspectives} could be a tangible step towards addressing 
the limitations of the current systems. 
Inherently this objective requires the understanding of the relations between each \perspective\, and \claim, as well as the nuance in semantic meaning between \perspective s under the context of the \claim. 

This work presents a demo for the task of  \emph{substantiated perspective discovery}~\cite{CKYCR19}. Our system receives a \emph{claim} and it is expected to present a \emph{diverse} set of \emph{well-corroborated} \emph{perspectives} that take a \emph{stance} with respect to the claim. Each perspective should be substantiated by \emph{evidence} paragraphs which summarize pertinent results and facts.

\begin{figure*}
    \centering
    \shadowbox{
    \includegraphics[scale=0.30,trim=0.6cm 0.0cm 0.6cm 0.3cm, clip=false]{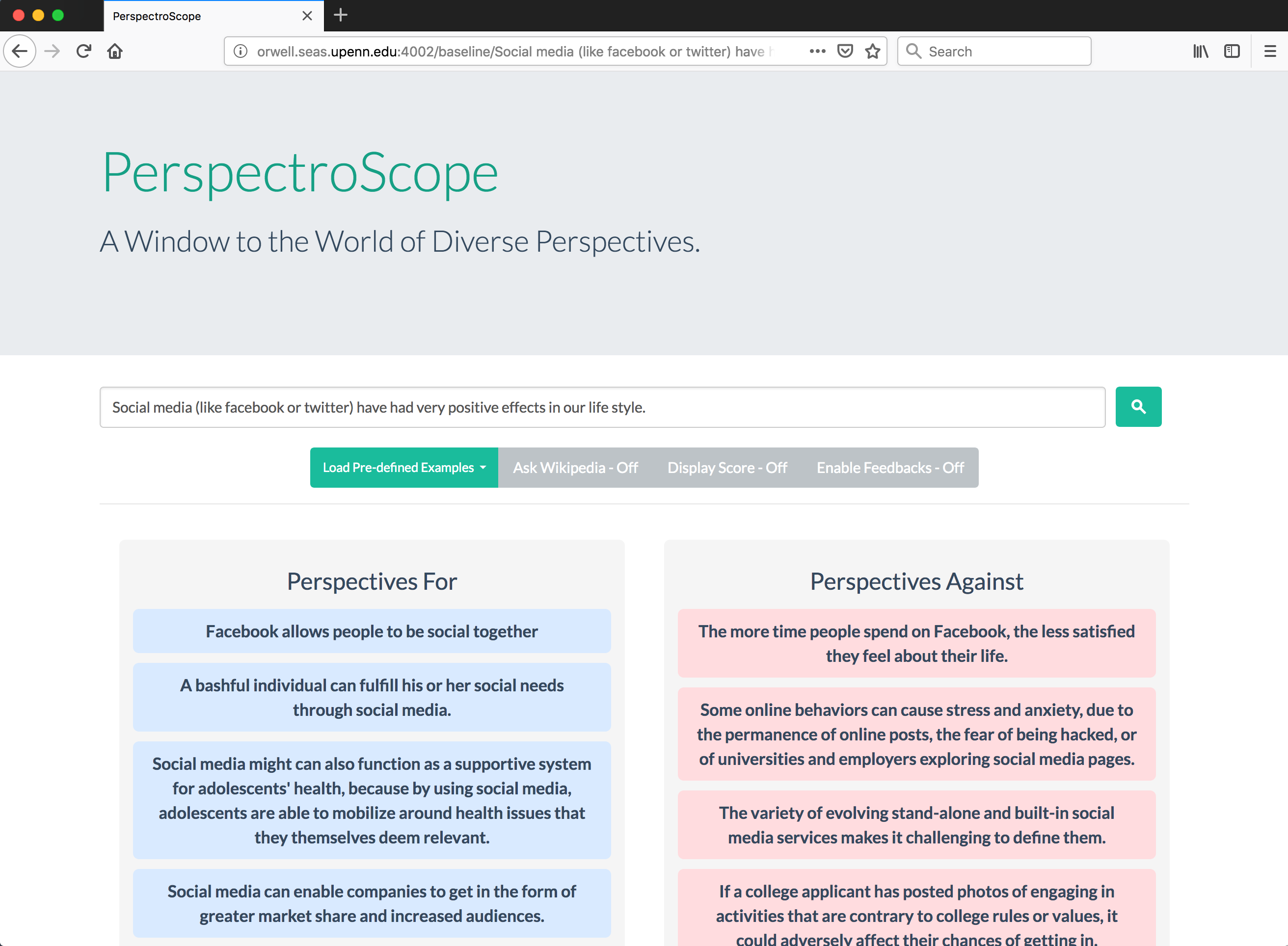}
    }
    \caption{
        Given a \emph{claim} as the input, our system is expected to discover various \emph{perspectives} and their \emph{stance} with respect to the claim. 
        Each claim also comes with the relevant \emph{evidence} that substantiates the given perspective.
    }
    \label{fig:example:intro}
\end{figure*}

A typical output of the system is shown in Figure~\ref{fig:example:intro}. The input to the system is a \emph{claim}: \emph{Social media (like facebook or twitter) have had very positive effects in our life style.} There is no single, best way to respond to the claim, but rather there are many valid responses that form a spectrum of perspectives, each with a {\em stance} relative to this claim and, ideally, with evidence supporting it.

To support the input claim, one could refer to the observation that interactions between individuals has become easier through the social media. Or one can refer to the success they have brought to those in need of reaching out to masses (e.g., business individuals). On the contrary, one could oppose the given claim by pointing out its negative impacts on productivity and the increase in cyber-bullying. 
Each of these arguments, which we refer to as a \perspective\ throughout the paper, is an opinion, possibly conditional, in support of a given \emph{claim} or against it.  
A \perspective\, thus constitutes a particular attitude towards a given \emph{claim}. 
Additionally, each of these \emph{perspective} has to be well-supported by \emph{evidence} found in  paragraphs that summarize findings and substantiations of different sources.

Overall, \name\ provides an interface to help individuals by providing a small but diverse set of \emph{perspectives}. Our system is built upon a few recent developments in the field. In addition, our system is designed to be able to utilize feedback from the users of the system to improve its predictions. 
The rest of this paper is dedicated to delineating the details of \name.

\section{\name}
\subsection{Core Design Structure}
\begin{figure*}
    \centering
    \includegraphics[scale=0.35]{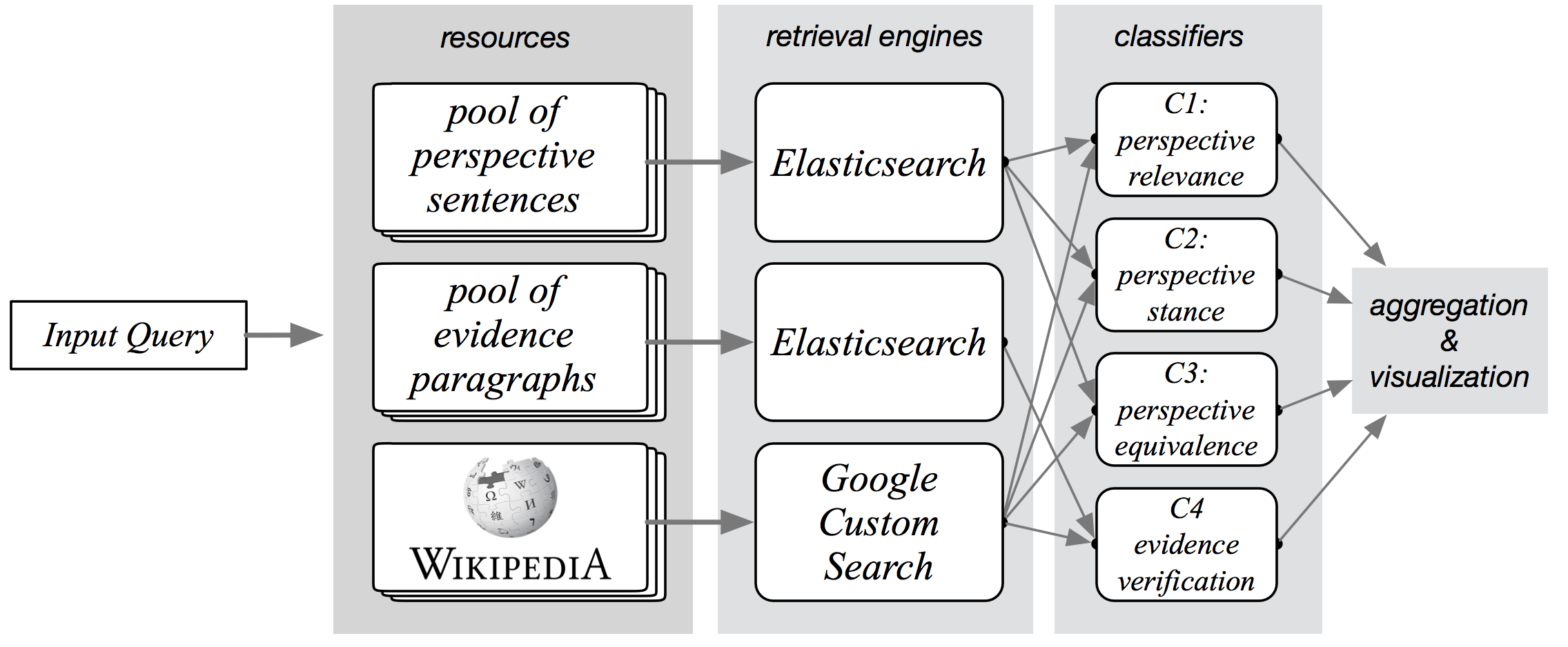}
    \caption{Overview of the system structure: given a query to the system, it extracts candidates from its internal knowledge  }
    \label{fig:overview}
\end{figure*}

A high-level picture of the work is shown in Figure~\ref{fig:overview}. 
Our system uses a mix of retrieval engines and learned classifiers to ensure both quality and efficiency. The retrieval systems extract candidates (perspectives or evidence paragraphs) which are later evaluated by carefully designed classifiers. 

\subsection{Learned Classifiers}
In building \name\ we borrow the definitions and dataset provided by \newcite{CKYCR19}.
The provided dataset, \datasetname, is a crowdsourced collection of claims, perspectives and evidence extracted from online debate websites as well as other web content. 
We follow the same steps as \newcite{CKYCR19} to create classifiers for the following tasks: 
\paragraph{C1: Relevant Perspective Extraction.} This classifier is expected to return the collection of perspectives with respect to a given claim.
\paragraph{C2: Perspective Stance Classification.} Given a claim, this classifier is expected to score a collection of perspectives with the degree to which it \emph{supports} or \emph{opposes} the given claim. 
\paragraph{C3: Perspective Equivalence.} This classifier is expected to decide whether two given perspectives are equivalent or not, in the context a given claim.  
\paragraph{C4: Extraction of Supporting Evidence.} This classifier decides whether a given document lends enough evidence for a given perspective to a claim.

In training the classifiers for each of the tasks, we use BERT~\cite{DCLT18} and we follow the same steps described in \newcite{CKYCR19}.

\begin{figure*}
    \centering
    \shadowbox{
    \includegraphics[scale=0.25,trim=0.6cm 0.0cm 0.6cm 0.3cm, clip=false]{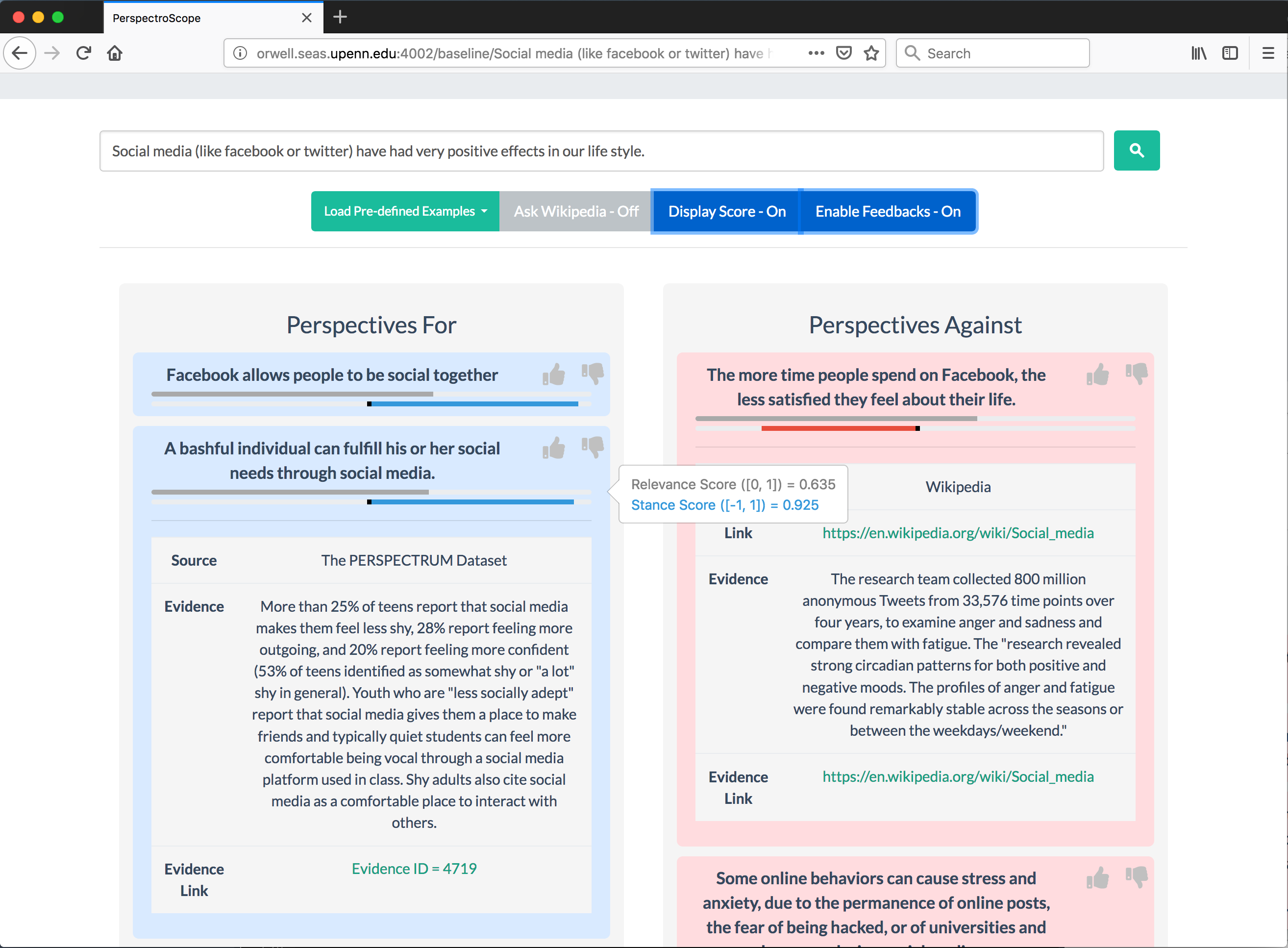}
    }
    \caption{
        A demonstration of the system features. The grey and blue/red color bars (under each perspective) show the relevance and stance predictions, respectively. Upon user request, the system provides a paragraph of supporting evidence for each perspective. Users have the option to provide feedback to each perspective via the \emph{thumbs-up} or \emph{thumbs-down} button.
    }
    \label{fig:example:intro}
\end{figure*}

\subsection{Candidate Retrieval}
We use a retrieval (IR) system\footnote{\url{www.elastic.co}} to generate \perspective\ and \evidence\ candidates for the learned classifiers. We take 10$k$ perspective sentences and 8$k$ evidence paragraphs from~\newcite{CKYCR19} and index them respectively in two independent retrieval engines. 
For each input claim, we query the claim and retrieve top-30 perspective candidates from the retrieval engine. 
Upon user request, we query the claim concatenated with a perspective candidate to retrieve top-20 evidence candidates from the pool of 8$k$ evidence paragraphs. 


To support a broader range of topics not covered by \datasetname, we use Wikipedia to retrieve extra candidate perspectives/evidence. Given an input claim from the user, we issue a query to the Google Custom Search API \footnote{\url{https://cse.google.com/cse/}} and retrieve top 10 relevant Wikipedia pages. We clean up each page using {\small \textsf{newspaper3k}}\footnote{\url{github.com/codelucas/newspaper}} and use the first sentence of the paragraphs within each document as candidate perspectives, and the rest sentences in each paragraph as candidate evidence.

\subsection{Minimal Perspective Discovery}
The overall decision making is outlined in Algorithm~\ref{alg:algorithm1}. 
As mentioned earlier, the whole process is a pipeline starting with candidate generation via retrieval engines, and followed by scoring with the learned classifiers. 
The final step is to select a \emph{minimal} set of perspectives with the \textsf{DBSCAN} clustering algorithm~\cite{ester1996density}.  

\begin{algorithm}
\SetAlgoLined
{
\small
\textbf{Input:} claim $c$. \\
\textbf{Output:} perspectives, their stances \& evidence. \\ 
$\hat{P} \leftarrow $IR$(c)$ \tcp{candidate perspectives } 
$P = \set{}$ \\ 
\ForEach{$p \in \hat{P}$}{
  \tcp{perspective relevance} 
  \If{ $C1(c, p) > T1$ and $abs(C2(c, p)) > T2$ }{ 
        $e \leftarrow C2(c, p)$\\ 
        $\hat{E} \leftarrow $IR$(c, p)$ \tcp{candidate evidence} 
        $E = \set{}$ \\ 
        \ForEach{$e \in \hat{E}$}{
            \tcp{evidence verification} 
            \If{ $C4(c, p, e) > T4$ }{ 
                $E \leftarrow E \cup \set{e}$. 
            }
        }
        $P \leftarrow P \cup \set{(p, s, E)}$. 
    }
  }
  $P \leftarrow $  \tcc{ minimal perspectives after clustering with \textsf{DBSCAN} on the equivalence scores between any perspective pairs via $C3$.  }
  }
 \caption{Minimal Perspective Extraction}
 \label{alg:algorithm1}
\end{algorithm}
The parameters of this algorithm (e.g., the thresholds $T1, T2, ...$) are tuned manually on a held-out set. 

\begin{figure*}
    \centering
    \shadowbox{\includegraphics[scale=0.39,trim=0cm 2cm 0.1cm 0.0cm, clip=true]{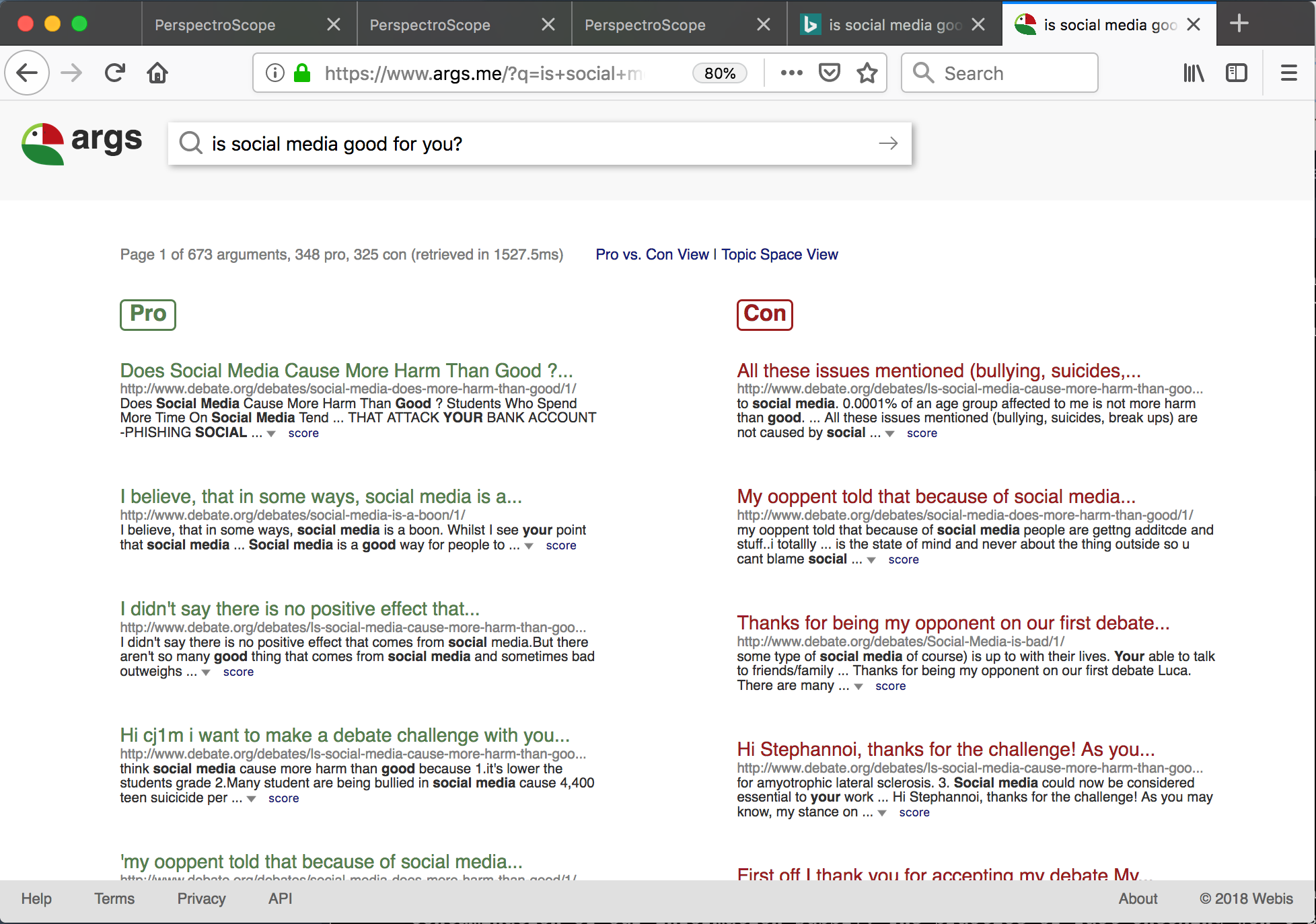}}
    \shadowbox{\includegraphics[scale=0.39,trim=0cm 0cm 0cm 0.1cm, clip=true]{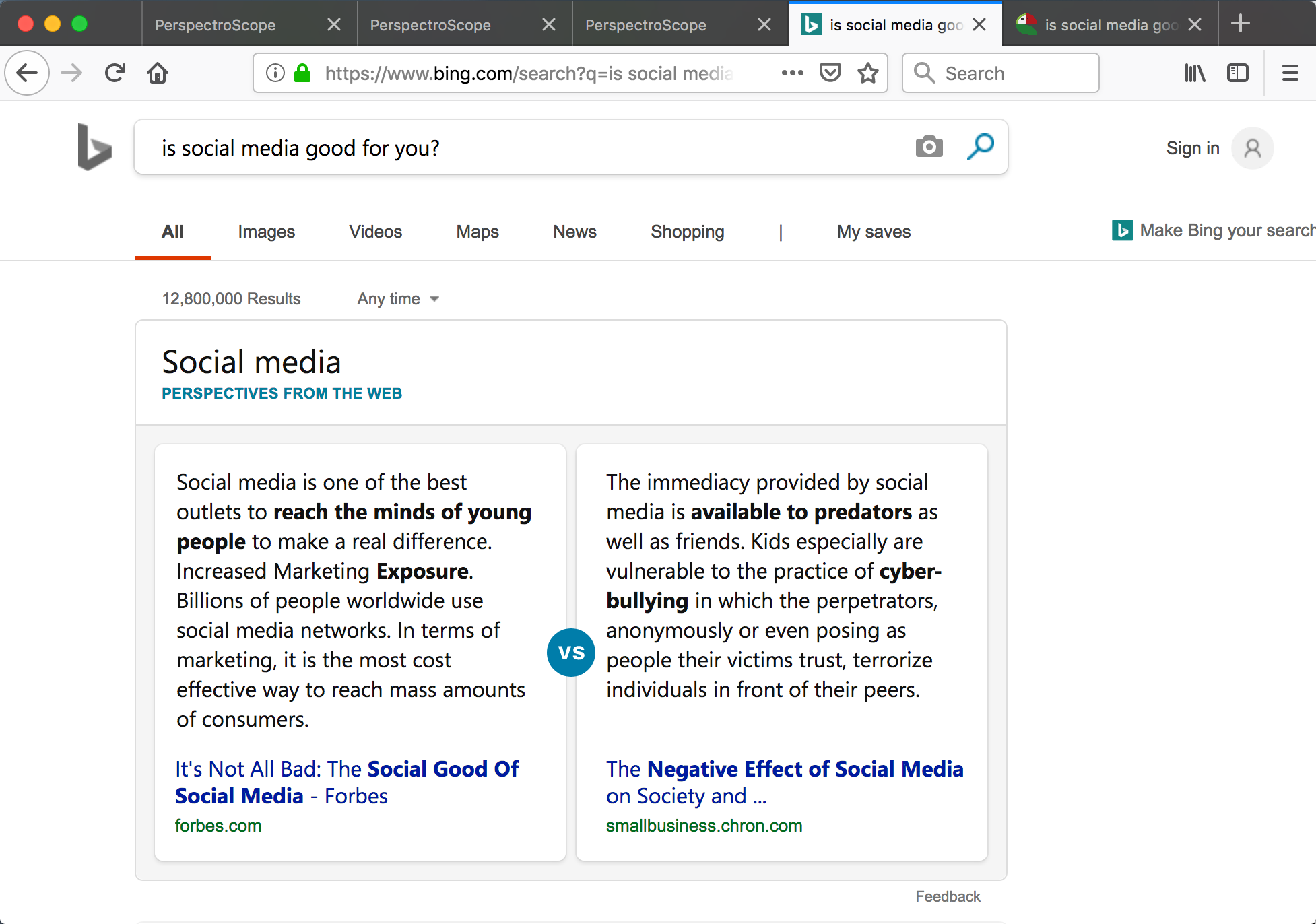}}
    \caption{Related work: {\tt args.me} an argument retrieval engine using arguments extracted from debate websites;  {\tt bing.com} search engine showing contrasting views on a debate topic.}
    \label{fig:others}
\end{figure*}

\subsection{Utilizing user feedback}
User feedback/logs are valuable sources of information for many successful applications. 
In this work, we collect two forms of feedback signals from users. We record all queries of claims issued to the system. In addition, the users have the option to tell us whether a given perspective is a good or bad one (based on the quality of its relevance, stance or evidence prediction). It is important to note that we are not collecting any personal information in the process.    

The user annotations can provide extra supervision signals for task C1-C4 with a broader topical coverage. These annotations can in turn be used in the classifier training and iteratively improve our prediction results with increasing number of users.  

\section{Related Work}

There are few related tools to this work.
{\tt args.me} is a platform that accepts natural language queries and returns links to the pages that contain relevant topics~\cite{wachsmuth2017building}, which are split into \emph{supporting} \& \emph{opposing}  categories (screenshot in Figure~\ref{fig:others}). 
Similarly, {\tt ArgumentText} \cite{stab2018argumentext} takes a topic as input and returns 
\emph{pro/con} arguments retrieved from the web. 
This work takes the effort one step further by employing language understanding techniques. 

There is a rich line of work on using Wikipedia as source for argument mining or to assess the veracity of a claim~\cite{TVCM18}. 
For instance, {\tt FAKTA} is a system that extracts relevant documents from Wikipedia, among other sources, to predict the factuality of an input claim \cite{nadeem2019fakta}.

Beyond published works, there are websites that employ similar technologies. For instance, {\tt bing.com} has recently started a service that provides two different responses to a given argument (screenshot in Figure~\ref{fig:others}). 
Since there is no published work on this system, it is not clear what the underlying mechanism is.


There exist a number of online debate platforms that provide similar functionalities as our system: 
{\tt \small  kialo.com, procon.org, idebate.org }, among others. 
Such websites usually provide a wide range of debate topics and various arguments in response to each topic. These resources have been proven useful in a line of works in argumentation \cite{HuaWa17, SMSRG18, wachsmuth2018retrieval}, among many others.  While they provide rich sources of information, their content is fairly limited in terms of  either their topical coverage or data availability for academic research purposes.

There also exist a few other works in this direction that do not accompany a publicly available tool or demo. For instance, \newcite{hasan2014you,LBGAS18} attempt to identify relevant arguments within web text in the context of a given topic. 

\section{Conclusion and Future Work}
We have presented \name, a powerful interface for exploring different perspectives to discussion-worthy claims. The system is built with a combination of retrieval engines and learned classifiers to create a good balance between speed and quality. Our system is designed with the mindset of being able to get feedback from users of the system.  

While this work offers a good step towards a higher quality and flexible interface, there are many issues and limitations that are not addressed here and are opportunities for future work. For instance, the system provided here does not provide any guarantees in terms of the  \emph{exhaustiveness} of the perspectives in the world, or levels of expertise and trustworthiness of the identified evidence. 
Moreover, any classifier trained on some annotated data (such as what we used here) could potentially contain hidden biases that might not be easy to see. 
We hope that some of these challenges and limitations will be addressed in future work.

\section*{Acknowledgments}
This work was supported in part by a gift from Google and by Contract HR0011-15-2-0025 with the US Defense Advanced Research Projects Agency (DARPA). The views expressed are those of the authors and do not reflect the official policy or position of the Department of Defense or the U.S. Government. 

\bibliography{main}
\bibliographystyle{acl_natbib}




\end{document}